\documentclass[11pt]{article}

\usepackage[final]{acl}
\usepackage{times}
\usepackage{latexsym}
\usepackage[T1]{fontenc}
\usepackage[utf8]{inputenc}
\usepackage{microtype}
\usepackage{inconsolata}
\usepackage{graphicx}
\usepackage[most]{tcolorbox}
\usepackage{nicematrix}
\usepackage{abbrevs}
\usepackage{xspace}
\usepackage{hyperref}
\usepackage{cleveref}
\usepackage{booktabs} 
\usepackage{bm}

\newcommand{\eg}{e.g.\xspace}
\newcommand{\ie}{i.e.\xspace}
\newcommand{\ourmethod}{\textsc{ViCE}\xspace}
\newcommand{\ourmethodlong}{\textbf{Vi}deo \textbf{C}aption \textbf{E}ntailment for video similarity estimation\xspace}
\newcommand{\ourtask}{ConViS\xspace}

\newcommand{\blind}{Blind \raisebox{-0.15\height}{\includegraphics[width=0.2cm]{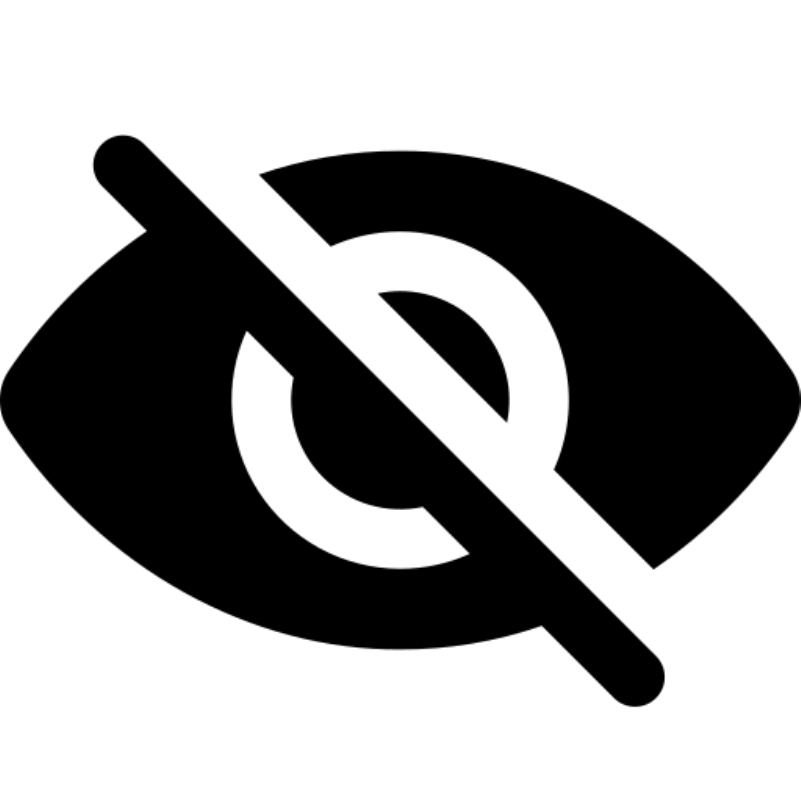}}}

\definecolor{goodgreen}{rgb}{0.0, 0.56, 0.0}
\definecolor{greengod}{RGB}{93, 184, 57}
\definecolor{badgray}{HTML}{666666}
\definecolor{badred}{RGB}{224, 53, 34}

\definecolor{sota}{HTML}{DFE8F6}
\definecolor{basemodel}{HTML}{F6F6F6}
\definecolor{header}{HTML}{EEEEEE}

\definecolor{tikzbaseline}{HTML}{FF2C71}
\definecolor{tikzours}{HTML}{027FFF}

\newcommand{\gooddelta}[1]{\scalebox{1.05}{$_{\textcolor{goodgreen}{\uparrow\bm{{#1}}}}$}}
\newcommand{\baddelta}[1]{\scalebox{1.05}{$_{\textcolor{badgray}{\downarrow\bm{{#1}}}}$}}

\newcommand{\plusours}{ \ \  \ourmethod}
\newcommand{\pluszs}{ \ \ \textit{zero-shot}}
\newcommand{\pluszscot}{ \ \  \textit{zero-shot CoT}}

\newtcolorbox{wideboxzero}[2][]{
    enhanced jigsaw,
    breakable,
    width=\textwidth,
    colback=white,
    title={#2},
    #1
}

\newtcolorbox{colbox}[2][]{
    enhanced jigsaw,
    breakable,
    colback=white,
    title={#2},
    #1
}

\newtcolorbox{widebox}[2][]{
    enhanced jigsaw,
    breakable,
    width=\textwidth,
    colback=white,
    title={#2},
    float*=ht,      
    #1
}

\definecolor{ZScolor}{HTML}{C19A7F}    
\definecolor{ViCEcolor}{HTML}{C73E55}  
\definecolor{GTcolor}{HTML}{A3B84C}

\title{Video-Language Entailment for Video Similarity Estimation}

\author{
 \textbf{Benedetta Liberatori\textsuperscript{1}},
 \textbf{Lorenzo Vaquero\textsuperscript{2}},
 \textbf{Elisa Ricci\textsuperscript{1,2}},
 \textbf{Paolo Rota\textsuperscript{1}}
\\
 \textsuperscript{1} University of Trento,
 \textsuperscript{2} Fondazione Bruno Kessler
\\
 \small{
   \textbf{Project website:} \href{https://benedettaliberatori.github.io/vice/}{https://benedettaliberatori.github.io/vice}
 }
}

\begin{document}
\maketitle
\begin{abstract}
Comparing videos to assess their semantic similarity remains a challenging task for current Large Multimodal Models (LMMs), which are primarily trained with single-video objectives. While most prior works have focused on computing global similarity, recent studies have introduced the more challenging problem of \textit{concept-based similarity}, which measures how similar two videos are with respect to specific concepts (\eg, activity, location). 
However, no dedicated approaches have yet leveraged LMMs to explicitly address this task. 
In this paper, we aim to fill this gap introducing \ourmethodlong (\ourmethod): a dual-branch design that leverages LMMs for cross-video reasoning.
Each branch jointly processes one video and the caption of the other, measuring their alignment for a specific concept via video-language entailment. 
The outputs from both branches are fused to produce a final similarity score, along with a textual explanation of the model reasoning. 
This approach leverages the model’s native single-video understanding while enabling fine-grained comparative reasoning between video and text. 
Experiments on ConViS-Bench and StepDiff show that \ourmethod outperforms models that process both videos simultaneously, highlighting the promise of single-video reasoning for concept-based similarity estimation.
\end{abstract}

\section{Introduction}
\label{sec:intro}

\begin{figure}[!t]
\includegraphics[width=0.9\linewidth]{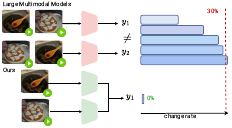}
\caption{
\textbf{Order sensitivity of Large Multimodal Models in video-pair understanding.}
Large Multimodal Models (LMMs) exhibit strong order dependence: swapping the order of videos in a pair leads to different predictions in approximately 25--30\% of cases. In contrast, our method is order-invariant (0\% change rate).}
\label{fig:teaser}
\end{figure}

Understanding video similarity lies at the core of numerous computer vision tasks, from video retrieval~\cite{wu2025learning} and content recommendation to reasoning over multi-source information~\cite{meng2025videoforest}.
Yet, defining what makes two videos \textit{similar} remains challenging, as similarity depends on several factors like semantics, context, and purpose. 
Two clips may depict an athlete kicking a ball in different environments, or people cooking pasta with different ingredients and utensils. 
Capturing semantic correspondences requires models that go beyond low-level visual cues and single-concept matches, revealing a deeper understanding of the events depicted in videos. 
This observation has motivated research on \textit{concept-based video similarity estimation} (\ourtask)~\cite{liberatori2025convisbench}: estimating how similar two videos are with respect to a specific concept (\eg, location, actions).
Unlike traditional global video similarity estimation~\cite{liu2024not,KordopatisZilos2023,VVSJo2024}, \ourtask requires models to selectively attend to one concept while ignoring irrelevant variations, making it more challenging.

Recent Large Multimodal Models (LMMs)~\cite{li2025llavaonevision, zhu2025internvl3, bai2025qwen2, zhang2025llavavideo} have achieved remarkable progress in video understanding. 
Currently, LMMs can accurately describe videos, answer questions about their content, and even make comparisons using natural language~\cite{nagarajan2024step, burgess2025video}.
Their generative nature allows them to be easily prompted with natural language, making them particularly suitable for \ourtask: by prompting them with a concept-specific instruction, users can query an LMM to estimate a video similarity score along the desired dimension, without the overhead of concept-specific fine-tuning. 
This aspect was recently investigated in~\cite{liberatori2025convisbench}, where the capabilities of different LMMs were analyzed and compared across a variety of concepts.

Previous works targeting video-pair understanding with LMMs feed both videos at once~\cite{zhu2025cvbench,liberatori2025convisbench,burgess2025video}. 
We empirically find that this approach suffers from a significant limitation: LMMs are highly sensitive to the ordering of the videos.
As illustrated in Fig.~\ref{fig:teaser}, simply reversing the order of videos in input can result in inconsistent judgments, with variations in responses ranging from $20\%$ to $30\%$ across different models.
These findings align with earlier observations in multiple-choice question answering, where the order of possible answers influences model predictions despite constituting semantically equivalent inputs~\cite{zheng2024large, alzahrani2024when, pezeshkpour2024large}. 
Such instability undermines the reliability of current LMMs for tasks that require robust similarity estimation between video pairs. 
We hypothesize that this brittleness stems  from a mismatch between their pretraining objectives and the requirements of the intended multi-video reasoning tasks.
Current LMMs are pretrained on single-video objectives (\eg, captioning or question answering)~\cite{bai2025qwen2}, limiting their ability to reason across multiple visual streams.
Consequently, multi-video inputs push these models outside their training regime, often yielding inconsistent outputs~\cite{meng2025videoforest, qiu2025videosetdiff, burgess2025video}.

We introduce \ourmethodlong (\ourmethod), a novel test-time inference approach that harnesses the inherent capabilities of LMMs while avoiding their weaknesses. 
Rather than feeding both videos simultaneously, we decompose the task into a dual-branch single-video pipeline: each branch evaluates entailment between one video and the textual description of the other with respect to a target concept, and the outputs are fused into a final similarity score with a textual explanation.
This design ensures LMMs operate strictly within their native single-video setting, avoiding degradation from multi-video prompts while enabling interpretable, language-grounded similarity judgments.

We evaluate our approach on ConViS-Bench~\cite{liberatori2025convisbench}, a benchmark explicitly designed for concept-conditioned video similarity. Additionally, we repurpose the StepDiff~\cite{nagarajan2024step} dataset, originally proposed for difference description in video pairs, by providing human-curated annotations following the same protocol as ConViS-Bench. 
Our evaluation, conducted on multiple LMMs, shows that \ourmethod consistently produces more semantically-faithful similarity estimates than directly inputting the video pairs. This shows that framing \ourtask within a caption-then-entail paradigm not only addresses the limitations of models trained with single-video objectives, but also improves interpretability and robustness in concept-based video comparison.

In summary, our contributions are threefold:
\textit{i)} We introduce \ourmethod, a dual-branch framework that leverages LMMs to compare videos using video-language entailment. To the best of our knowledge, it is the first method specifically designed for this task.
\textit{ii)} \ourmethod operates within the model's native single-video setting, yielding more semantically faithful similarity estimates and interpretable textual explanations.
\textit{iii)} We validate \ourmethod on ConViS-Bench and repurpose the StepDiff dataset with concept-level human similarity judgments, highlighting the effectiveness of caption-then-entail reasoning on multiple LMMs.
\section{Related Work}
\label{sec:related}

\begin{figure*}[t]
    \centering
    \includegraphics[width=0.9\linewidth]{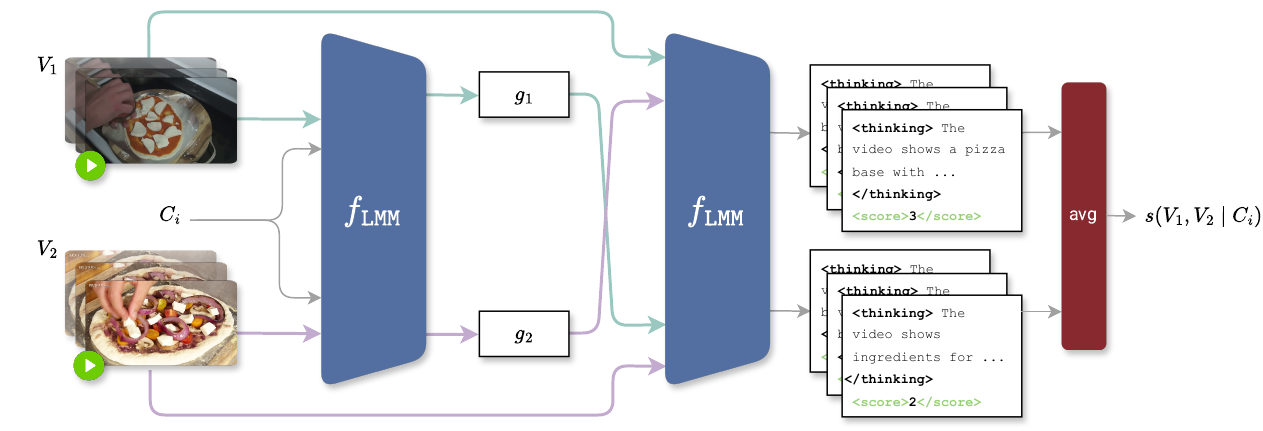}
    \caption{\textbf{Overview of \ourmethod:} Given a video pair $(V_1, V_2)$ and a concept $C_i$, \eg, \textbf{\texttt{Ingredients}}, \ourmethod processes the videos through two parallel branches. Each branch generates a concept-conditioned caption $(g_1, g_2)$, then performs video–caption entailment by comparing one video with the caption of its counterpart. The branches produce reasoning traces inside \textbf{\texttt{<thinking>\ldots</thinking>}} and similarity scores inside \textbf{\texttt{<score>\ldots</score>}}. The final score $s(V_1,V_2\mid C_i)$ is averaged across both branches. For clarity, we omit visualizations of the reasoning trace summarization (Sec.~\ref{sec:explain}), as it does not affect the predicted score and is used solely for interpretability.}
    \label{fig:method}
\end{figure*}

\noindent\textbf{Large Multimodal Models for Video.}
Recent advances in foundation models have led to the emergence of architectures capable of processing videos natively~\cite{zhang2025llavavideo}.
Meanwhile, several works have extended image-centric models to the video domain, allowing unified understanding across static and dynamic visual data~\cite{ye2025mplugowl,zhu2025internvl3,wang2025qwen2,bai2025qwen2,bai2025qwen3}.
Despite these developments, most existing LMMs remain limited in their ability to jointly process and reason over multiple distinct videos.
Recent efforts have started to investigate large-scale video foundation models in multi-video contexts~\cite{meng2025videoforest,zhu2025cvbench,qiu2025videosetdiff}, uncovering significant limitations in the current architectures.
These models are neither architecturally designed nor pretrained to jointly reason over multiple videos, limiting their effectiveness in tasks that require integrated understanding across distinct visual streams.
Differently, this work specifically focuses on reasoning over two videos, as required by video similarity estimation.

\noindent\textbf{Concept-based Video Similarity Estimation.}
Motivated by prior work on concept-based image similarity~\cite{achille2024interpretable}, recent research has explored how to compare videos with respect to specific semantic concepts.
VidDiffBench~\cite{burgess2025video} advances this direction by prompting models to generate textual explanations of subtle variations in how actions are performed, focusing on fine-grained differences.
StepDiff~\cite{nagarajan2024step}, in contrast, investigates difference-centric descriptions of procedural videos and extends this paradigm by comparing videos along multiple dimensions relevant to the domain, such as ingredients, tools, or techniques.
However, its scope remains largely limited to cooking-related content, emphasizing difference captioning within this narrow procedural domain.
To address this, ConViS-Bench~\cite{liberatori2025convisbench} introduces quantitative, concept-specific similarity measures across a broader spectrum of domains and semantic categories (\eg, objects, locations, actions).
While ConViS-Bench focuses on benchmarking LMMs, our work introduces the first dedicated approach for concept-based video similarity estimation.

\noindent\textbf{Video-Language Entailment.}
Video-language entailment examines whether a video semantically supports a given textual hypothesis.
Originally posed as a binary classification task~\cite{liu2020violin}, it has evolved into a framework for fine-grained reasoning and compositional inference~\cite{saravanan2025velociti}, with early works exploring multiple pretraining objectives and explicit visual grounding~\cite{li2020hero, chen2021explainable, sanders2024tvtrees}.
Recent work leverages LMMs for entailment directly through prompting~\cite{zhao2025can}.
Advances along this line include task-specific fine-tuning with synthetic videos~\cite{zanella2025can} and the use of textual explanations to guide model decisions and enhance interpretability~\cite{bansal2024videocon}.
In this work, we extend video-language entailment to multi-video reasoning.

\section{Method}
\label{sec:method}

We introduce \ourmethodlong (\ourmethod), a test-time inference, dual-branch method that uses LMMs to estimate concept-conditioned similarities between video pairs.
In the following, we first formally define the 
task (Sec.~\ref{sec:prob_form}), then describe our proposed \ourmethod (Sec.~\ref{sec:vice}), and show how we can synthesize human-readable explanations from it (Sec.~\ref{sec:explain}).

\subsection{Problem Formulation}
\label{sec:prob_form}

Given a video pair and a concept of interest, \textit{concept-based video similarity estimation} (\ourtask) aims to quantify how similar the two videos are with respect to that specific concept. 
Formally, let $\mathcal{C}=\{C_1,\dots,C_K\}$ denote a set of concepts in natural language, and $\left(V_1, V_2\right)$ a video pair. \ourtask aims to predict a concept-conditioned similarity score $s(V_1,V_2 \mid C_i) \in [1,5]$, for each $C_i\in\mathcal{C}$.

For example, consider the two cooking videos shown in Fig.~\ref{fig:method}.
Conditioned on $C_i=\textit{``action''}$, the similarity should be high, as both videos depict adding toppings to a pizza.
Conditioned on $C_i=\textit{``ingredients''}$, it should be lower, as the videos differ in toppings and cheese type.

\subsection{Video Caption Entailment for ConViS}
\label{sec:vice}

Large multimodal models have demonstrated impressive zero-shot capabilities on many downstream video tasks, primarily driven by large-scale pretraining on video-text data~\cite{zhang2025llavavideo,zhu2025internvl3,bai2025qwen2}.
However, these models remain much stronger at processing individual videos than reasoning across multiple video inputs simultaneously~\cite{meng2025videoforest,zhu2025cvbench,qiu2025videosetdiff}.

Building upon these findings, we propose \ourmethod, a zero-shot framework that leverages LMMs to estimate concept-based similarity between video pairs.
As illustrated in Fig.~\ref{fig:method}, \ourmethod follows a two-stage pipeline with two parallel branches. The first stage, \textit{concept-conditioned captioning}, generates textual descriptions for each video that explicitly emphasize the target concept.
In the second stage, \textit{video-caption entailment}, each video is paired with the other video's caption and fed to an LMM, which evaluates how well the video content aligns with the corresponding concept-conditioned description. 

\noindent\textbf{Concept-conditioned captioning.} Let $f_{\texttt{LMM}}$ denote the LMM and $j \in \{1,2\}$ the index of the branch. 
In the first stage, each branch produces a \emph{concept-conditioned} textual description of each video in the pair $\left(V_1, V_2\right)$,  explicitly reflecting the conditioning concept: 
\begin{equation}
    f_{\texttt{LMM}}(V_j, C_i) = g_j, \quad j \in \{1, 2\}, \; C_i \in \mathcal{C},
\label{eq:caption_gen}
\end{equation}
where $g_j$ represents the concept-conditioned caption of video $V_j$ produced by branch $j$ under conditioning $C_i$.

In this phase, we propose to leverage the zero-shot capabilities of the LMM, which was pre-trained on single-video captioning tasks~\cite{zhang2025llavavideo,zhu2025internvl3,bai2025qwen2}, allowing it to generate meaningful concept-conditioned captions from a single video in input.
Back to our example in Fig.~\ref{fig:method}, conditioned on the concept $C_i=\textit{``ingredients''}$, the model produces captions such as \textit{``tomato sauce, mozzarella cheese, and basil''} for $V_1$ and \textit{``mushrooms, tomatoes, red onion, cheese, and herbs''} for $V_2$.  
In contrast, for $C_i=\textit{``action''}$, both videos are described as \textit{``adding toppings to the pizza''}, highlighting the shared procedural aspect. 

\noindent\textbf{Video-caption entailment.} In the second stage, \ourmethod measures semantic alignment between the two videos by evaluating bidirectional entailment (\ie, how well each video supports the other's concept-conditioned caption).
For videos $V_1$ and $V_2$, this is performed in both directions:
\begin{equation}
    f_{\texttt{LMM}}(V_j, g_k, C_i) = \tau_j, j,k \in \{1,2\}, \; j \neq k,
\label{eq:ent_step}
\end{equation}
where $\tau_j$ is the entailment output of the model.  

Judging the degree to which a video entails a description requires complex reasoning with at least two steps: understanding the content of the video and assessing the relevance of the description. 
To stimulate this reasoning capability, we generate the output $\tau_j$ as a sequence of intermediate steps, leading to the final entailment score $a_j$ as: 
\begin{equation}
\label{eq:traces_with_score}
    \tau_j = [z_{j_1}, \dots, z_{j_N}, a_j] \sim p(\cdot \mid V_j, g_k, C_i),
\end{equation} 
where $p(\cdot)$ is the probability distribution over the vocabulary of $f_{\texttt{LMM}}$ and each intermediate step is generated autoregressively as:
\begin{equation}
    z_{j_n} \sim p( \cdot \mid V_j, g_k, C_i, z_{j_1}, \dots, z_{j_{n-1}}),
\end{equation}
for $n = 1, \dots, N, \; j,k \in \{1,2\}, \; j \neq k$.

With reference to Fig.~\ref{fig:method}, intermediate steps $z_{j_n}$ and final score $a_j$ are wrapped in \texttt{<thinking>}\ldots\texttt{</thinking>} and \texttt{<score>}\ldots\texttt{</score>} tags, constituting a reasoning trace.
This syntax not only provides free-form rationales describing the evidence and comparisons relevant to the concept $C_i$, but also actively guides the model's decision process~\cite{kojima2022large}.

\begin{table*}[ht!]
\def\arraystretch{1.0}
\setlength{\tabcolsep}{2pt}
\centering
\scriptsize
\resizebox{\textwidth}{!}{%
\begin{NiceTabular}{l@{\hskip29pt}c@{\hskip20pt}c@{\hskip19pt}c@{\hskip30pt}c@{\hskip19pt}c@{\hskip18pt}c@{\hskip 0pt}l}
\CodeBefore
   \rowcolors{5-6, 8-9, 13-14, 17-18}{sota}{}
   \rowcolors{2-3, 5-6, 9-10, 14-15, 18-19}{}{basemodel}
   \rowcolors{20-21}{sota}{}
\Body
\toprule
Model & 
\textbf{\texttt{Main Action}} &
\textbf{\texttt{Main Subjects}} &
\textbf{\texttt{Main Objects}} &
\textbf{\texttt{Location}} &
\textbf{\texttt{Order of Actions}} &
\multicolumn{2}{l}{Avg.} \\
\midrule
LLaVA-OV-7B~\cite{li2025llavaonevision} & & & & & & \\
\pluszs & 51.8 & 48.4 & \textbf{58.6} & 58.9 & \textbf{41.0} & \textbf{51.7} \\
\pluszscot & 39.9 & 28.6 & 28.3 & 44.1 & 28.0 & 33.8 \\
\plusours & \textbf{54.4} & \textbf{55.9} & 50.0 & \textbf{64.8} & 19.2 & 48.9 & \baddelta{2.8} \\
\midrule
\multicolumn{7}{l}{LLaVA-Video-7B~\cite{zhang2025llavavideo}} \\
\pluszs & 44.2 & 39.8 & 45.8 & 56.0 & 41.2 & 45.4 \\
\pluszscot & 25.1 & 25.1 & 31.3 & 30.2 & 14.7 & 25.3 \\
\plusours & \textbf{63.8} & \textbf{56.0} & \textbf{60.1} & \textbf{64.7} & \textbf{57.9} & \textbf{60.5} & \gooddelta{15.1}\\
\midrule
\multicolumn{7}{l}{Qwen3-VL-2B~\cite{bai2025qwen3}} \\
\pluszs     & 29.2 & 31.3 & 35.9 & 40.9 & 16.6 & 30.8 \\
\pluszscot  & 31.5 & 37.5 & 38.2 & 59.1 & 9.8  & 35.2 \\
\plusours   & \textbf{58.4} & \textbf{50.7} & \textbf{60.0} & \textbf{61.8} & \textbf{52.3} & \textbf{56.3} & \gooddelta{25.5} \\

\midrule
\multicolumn{7}{l}{Qwen3-VL-8B~\cite{bai2025qwen3}} \\
\pluszs     & 43.0 & 43.3 & 47.1 & 52.8 & 35.7 & 44.4 \\
\pluszscot  & 51.0 & 46.9 & 50.2 & 64.5 & 41.1 & 50.7 \\
\plusours   & \textbf{64.6} & \textbf{49.6} & \textbf{61.4} & \textbf{69.8} & \textbf{62.7} & \textbf{61.6} & \gooddelta{17.2} \\

\midrule
\multicolumn{7}{l}{Qwen3-VL-32B~\cite{bai2025qwen3}} \\
\pluszs     & 60.4 & 50.2 & 51.8 & 58.8 & 52.7 & 54.8 \\
\pluszscot  & 65.3 & 52.8 & 57.7 & 70.1 & 64.5 & 62.1 \\
\plusours   & \textbf{69.4} & \textbf{61.7} & \textbf{61.4} & \textbf{70.2} & \textbf{65.3} & \textbf{65.6} & \gooddelta{10.8} \\
\end{NiceTabular}
}
\caption{\textbf{Correlation performance on ConViS-Bench}~\cite{liberatori2025convisbench}. We apply \ourmethod to multiple LMMs, reporting Spearman’s $\rho$ correlations with human judgment ($\times100$). 
}
\label{tab:convs_main}
\end{table*}

Returning to our example in Fig.~\ref{fig:method}, when evaluating whether $V_1$ entails $V_2$'s caption for $C_i=\textit{``ingredients''}$, the reasoning trace might focus first on observing \textit{mushrooms} in $V_1$, then comparing to $V_2$'s mention of \textit{tomato sauce, mozzarella, and basil}, and finally output a lower entailment score for the \textit{ingredients} concept.  
For the $C_i=\textit{``action''}$ concept, the two directions observe the videos adding toppings and output a high entailment score, reflecting the common procedure.

To improve robustness, we generate reasoning traces $\tau_j^{m}$ through repeated sampling from $f_{\texttt{LMM}}$:
\begin{equation}
\{\tau_j^{1}, \dots, \tau_j^{M}\} \quad \text{for } m = 1,\dots,M.
\end{equation}
These reasoning traces may capture a diverse set of scores  
and mitigate the impact of outlier predictions on the semantic alignment evaluation~\cite{wang2023selfconsistency}, as sampled traces can lead to different final entailment scores.

Lastly, the final concept-conditioned similarity score between $V_1$ and $V_2$ under concept $C_i$ is obtained by averaging the entailment scores across all generated traces, as in~\cref{eq:traces_with_score}, and both branches:
\begin{equation}
    s(V_1, V_2 \mid C_i) = \frac{1}{2M} \sum_{j=1}^{2} \sum_{m=1}^{M} a_j^{m},
\end{equation}
where $a_j^{m}$ is the score from the $m$-th trace $\tau_j$.

\noindent\textbf{Asymmetry between branches.}
Importantly, we do not require the two branches to produce mutually consistent scores. 
Because entailment is inherently asymmetric, it is possible that the caption $g_1$ distilled from $V_1$ describes only a subset of the ingredients that appear in $V_2$, while $g_2$ distilled from $V_2$ lists a richer set of toppings. 
In this case, the entailment score for $(V_2, g_1, C_i)$ can be high, as $V_2$ easily satisfies the simpler description, whereas the reverse direction $(V_1, g_2, C_i)$ may yield a lower score, since $V_1$ does not contain all the ingredients referenced in $g_2$. 
We deliberately preserve this asymmetry, as it captures whether one video conceptually subsumes the other under $C_i$; by combining the scores from both branches, \ourmethod arrives at a more reliable and balanced similarity estimate.

\subsection{From Reasoning Traces to Explanations}
\label{sec:explain}

While \ourmethod aggregates entailment scores across directions and reasoning traces, its intermediate \texttt{<thinking>}\ldots\texttt{</thinking>} traces also contain rich, concept-specific evidence (\eg, object mentions, temporal cues, counterexamples).
To better understand the underlying $f_{\texttt{LMM}}$ model's behavior and make this information useful for real-world applications, we synthesize a single, human-readable explanation by summarizing the reasoning traces generated for both directions. 
Let:
\begin{equation}
\mathcal{R}_i = \big\{\mathbf{z}_{j}^{m}\ \big|\ j\in\{1,2\},\, m\in\{1,\dots,M\}\big\}
\end{equation}
denote the multiset of reasoning traces associated with concept $C_i$, where each $\mathbf{z}_{j}^{m}=[z_{j_1}^m, \dots, z_{j_N}^m]$ denotes the generated reasoning trace omitting the final score.
We employ a large language model $f_{\texttt{LLM}}$ to produce a concept-grounded explanation:
\begin{equation}
\label{eq:sum}
e_i \;=\; f_{\texttt{LLM}} \big(\mathcal{R}_i\big).
\end{equation}
This explanation covers consensus evidence supporting similarity under $C_i$.
This step is strictly post hoc and does not modify the numeric similarity $s(V_1,V_2\mid C_i)$.
It yields a holistic, concept-aligned narrative that explains the model's decision, making \ourmethod transparent and practical for downstream use.
We validate the quality of these explanations through a human evaluation study; details and results are provided in Appendix~\ref{sec:user_study}.
\section{Experiments}
\label{sec:experiments}

In this section, we present the experimental protocol and evaluate \ourmethod on the \ourtask task, showing improvements across multiple LMMs through extensive quantitative and qualitative analyses. 

\noindent\textbf{Dataset and Metrics.}
We use two datasets designed for \ourtask: ConViS-Bench~\cite{liberatori2025convisbench} and StepDiff~\cite{nagarajan2024step}. 
ConViS-Bench comprises 610 video pairs, annotated with human judgments across five broad concepts: main action, main subjects, main objects, location, and order of actions. 
StepDiff contains 5946 test videos, originally annotated on a subset of five cooking-related dimensions: technique, tools/equipment, actions, visuals, and ingredients.
Since the public release included only partial annotations, we re-annotated the full set with human annotators to ensure complete and consistent coverage on 1269 video pairs. 
Details on the annotation process are provided in Appendix~\ref{sec:stepdiff}.
We use Spearman's $\rho$ (scaled by $\times100$) to evaluate correlation with humans, following standard protocols~\cite{achille2024interpretable}, as it provides a robust measure in the presence of ties~\cite{puth2015effective}. 
Implementation details and prompts are provided in Appendix~\ref{sec:impl_det} and~\ref{sec:prompts}, respectively.

\begin{table*}[ht!]
\def\arraystretch{1.0}
\setlength{\tabcolsep}{1pt}
\centering
\scriptsize
\resizebox{\textwidth}{!}{%
\begin{NiceTabular}{l@{\hskip32pt}c@{\hskip12pt}c@{\hskip15pt}c@{\hskip30pt}c@{\hskip25pt}c@{\hskip25pt}c@{\hskip 0pt}l}
\CodeBefore
   \rowcolors{5-6, 8-9, 13-14, 17-18}{sota}{}
   \rowcolors{2-3, 5-6, 9-10, 14-15, 18-19}{}{basemodel}
   \rowcolors{20-21}{sota}{}
\Body
Model & \textbf{\texttt{Technique}} & \textbf{\texttt{Tools/Equipment}} & \textbf{\texttt{Actions}} & \textbf{\texttt{Visuals}} & \textbf{\texttt{Ingredients}} & Avg. \\
\midrule
\multicolumn{7}{l}{\textit{LLaVA-OV-7B}~\cite{li2025llavaonevision}} \\
\pluszs & 31.1 & \textbf{39.4} & 32.8 & \textbf{32.3} & \textbf{28.5} & 32.8 \\
\pluszscot & 29.5 & 35.3 & 29.4 & 25.3 & 24.1 & 28.7 \\
\plusours & \textbf{42.2} & 36.6 & \textbf{32.8} & 31.8 & 26.5 & \textbf{34.0} & \gooddelta{1.2} \\
\midrule
\multicolumn{7}{l}{\textit{LLaVA-Video-7B}~\cite{zhang2025llavavideo}} \\
\pluszs & 30.6 & 37.4 & 28.9 & \textbf{29.2} & 17.5 & 28.7 \\
\pluszscot & 26.3 & 35.3 & 27.4 & 22.7 & 15.1 & 25.4\\
\plusours & \textbf{48.4} & \textbf{37.8} & \textbf{34.4} & 28.5 & \textbf{34.5} & \textbf{36.7} & \gooddelta{8.0} \\
\midrule
\multicolumn{7}{l}{Qwen3-VL-2B~\cite{bai2025qwen3}} \\
\pluszs & 11.7 & 20.2 & 12.9 & 5.5 & 18.5 & 13.8 \\
\pluszscot & 18.5 & 25.6 & 20.3 & 15.3 & 14.7 & 18.9 \\
\plusours & \textbf{48.1} & \textbf{47.1} & \textbf{50.2} & \textbf{39.8} & \textbf{25.9} & \textbf{42.2} & \gooddelta{23.3} \\
\midrule
\multicolumn{7}{l}{Qwen3-VL-8B~\cite{bai2025qwen3}} \\
\pluszs & 42.6 & \textbf{52.6} & 48.2 & 41.8 & 35.9 & 44.2 \\
\pluszscot & 43.4 & 47.4 & 47.0 & 41.3 & 36.7 & 43.2 \\
\plusours & \textbf{53.7} & 50.1 & \textbf{49.2} & \textbf{43.0} & \textbf{39.8} & \textbf{47.2} & \gooddelta{3.0} \\

\midrule
\multicolumn{7}{l}{Qwen3-VL-32B~\cite{bai2025qwen3}} \\
\pluszs     & 47.1 & 51.3 & 49.2 & 42.1 & 43.5 & 46.6 \\
\pluszscot  & 48.2 & 50.0 & 51.6 & 43.8 & 45.4 & 47.8 \\
\plusours   & \textbf{50.1} & \textbf{52.2} & \textbf{51.8} & 39.8 & \textbf{45.4} & \textbf{47.9} & \gooddelta{1.3} \\
\end{NiceTabular}
}
\caption{\textbf{Correlation performance on StepDiff}~\cite{nagarajan2024step}.  We apply \ourmethod to multiple LMMs, reporting Spearman’s $\rho$ correlations with human judgment ($\times100$). 
}
\label{tab:stepdiff_main}
\end{table*}

\subsection{Main Results}

We evaluate our proposed \ourmethod on multiple strong off-the-shelf LMMs for video understanding: Qwen3-VL-2B, Qwen3-VL-8B, Qwen3-VL-32B~\cite{bai2025qwen3}, LLaVA-OneVision-7B~\cite{li2025llavaonevision}, LLaVA-Video-7B~\cite{zhang2025llavavideo}. 
As no prior work has addressed \ourtask with a dedicated methodology, we adopt these LMMs as baselines.
We employ a \textit{zero-shot} baseline where the two videos are concatenated into a single input, following prior video-pair understanding works~\cite{liberatori2025convisbench}.
We further extend this baseline with a chain-of-thought variant (\ie, \textit{zero-shot CoT}), prompting the model to produce an explicit reasoning trace before the final output score.

\begin{figure}[]
    \centering

    \begin{minipage}[t]{\linewidth}
        \centering
        \includegraphics[width=\linewidth]{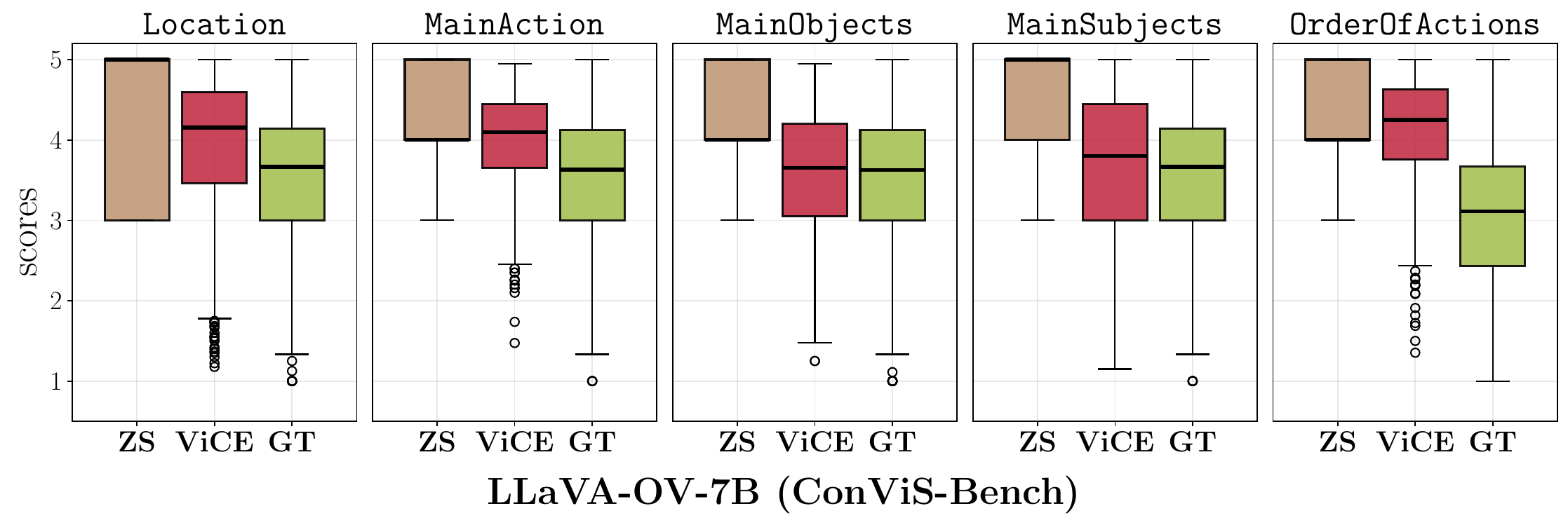}
    \end{minipage}
    \begin{minipage}[t]{\linewidth}
        \centering
        \includegraphics[width=\linewidth]{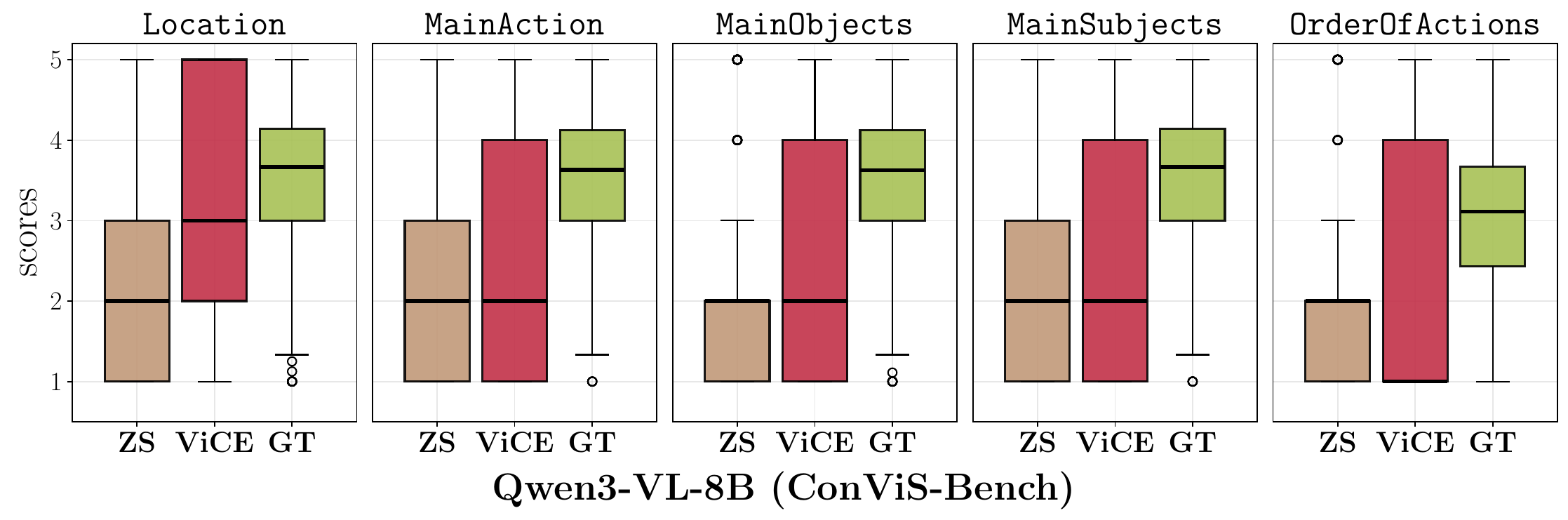}
    \end{minipage}

    \begin{minipage}[t]{\linewidth}
        \centering
        \includegraphics[width=\linewidth]{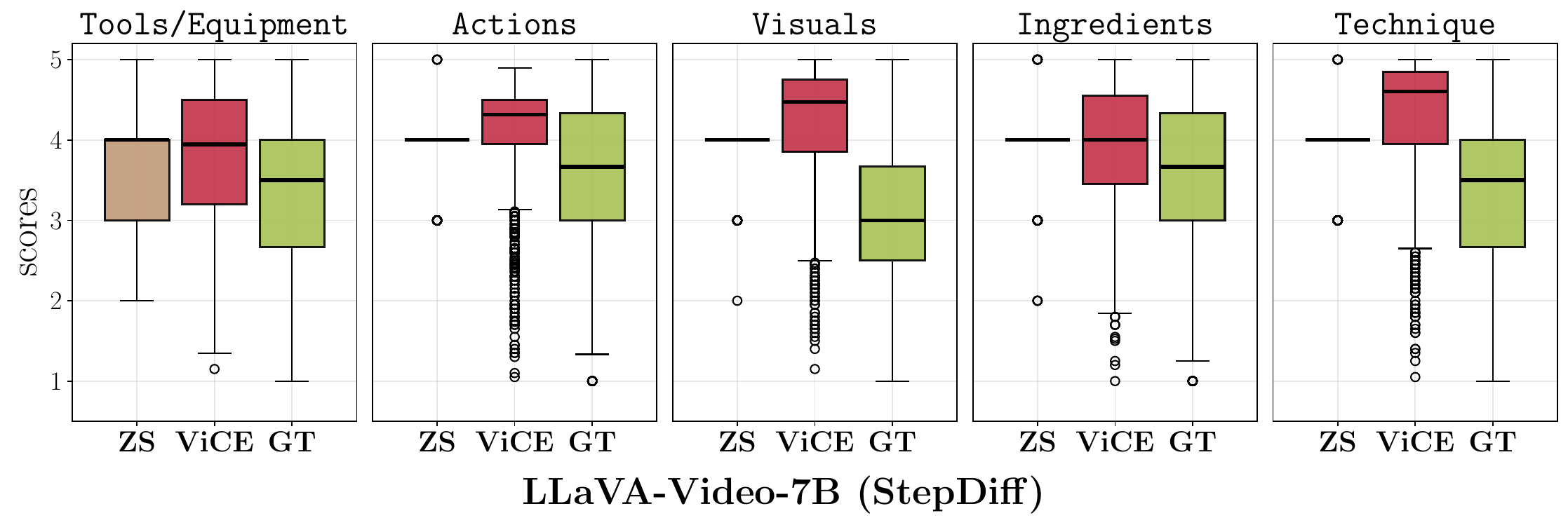}
    \end{minipage}
    \begin{minipage}[t]{\linewidth}
        \centering
        \includegraphics[width=\linewidth]{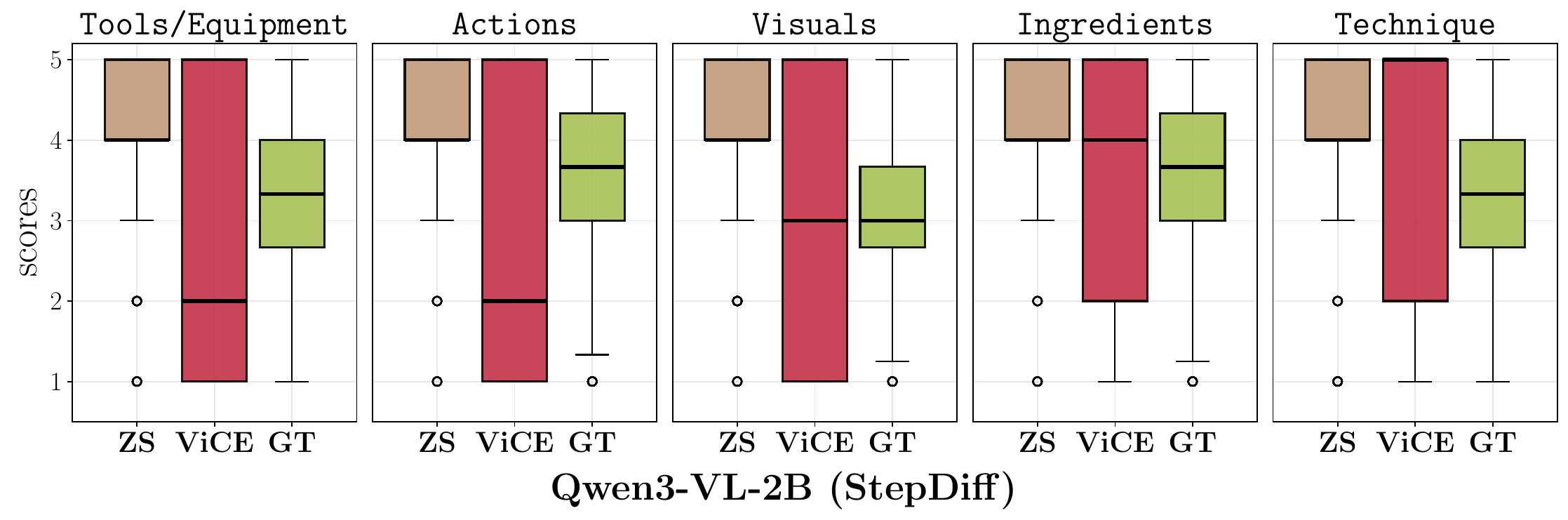}
    \end{minipage}

    \caption{
    \textbf{Distributions of predicted scores across datasets and models.}
    We compare \textcolor{ZScolor}{zero-shot} predictions, predictions with \textcolor{ViCEcolor}{\ourmethod}, and \textcolor{GTcolor}{ground truths}.
    }
\vspace{-10pt}
    \label{fig:box_plots}
\end{figure}

\begin{figure}[t]
    \centering
    \includegraphics[width=\linewidth]{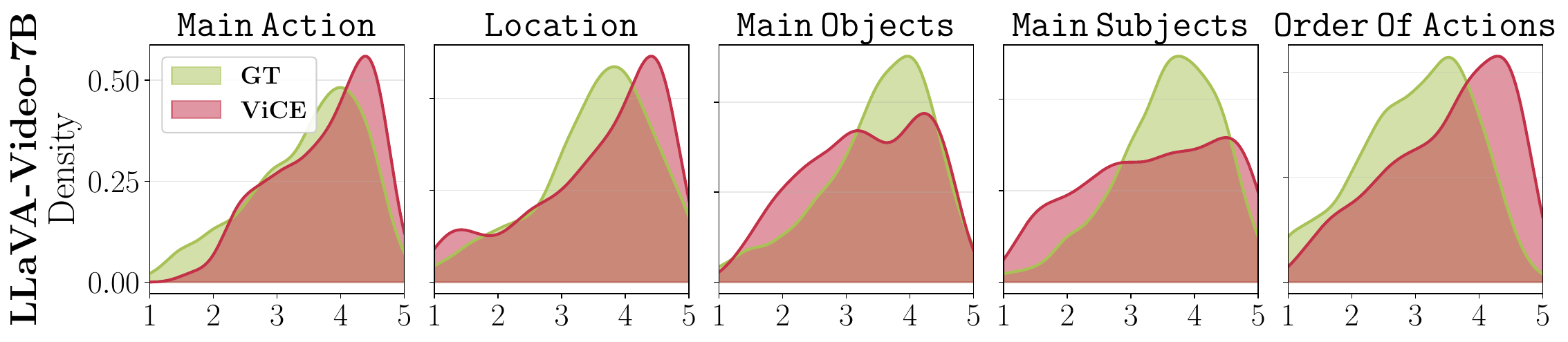} \\
    \includegraphics[width=\linewidth]{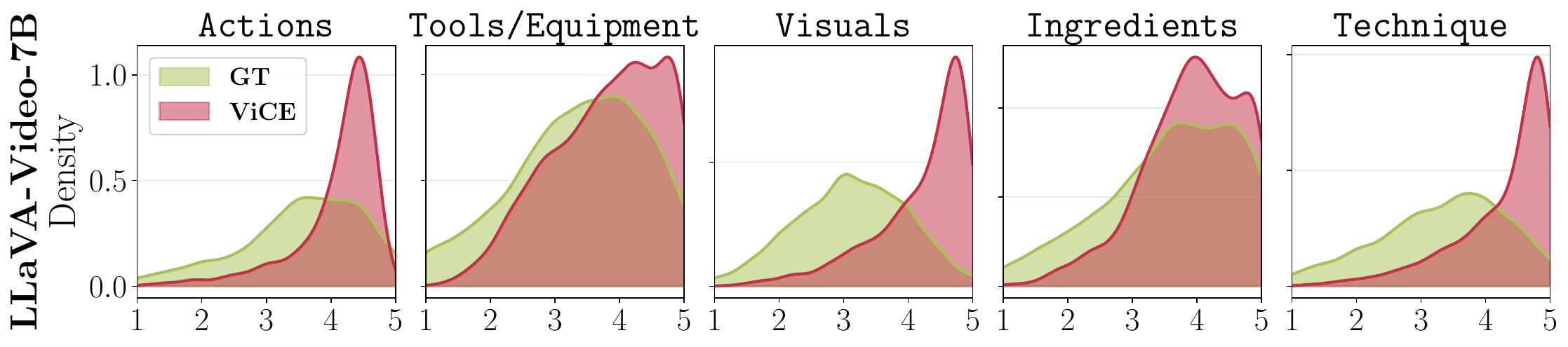} \\
    \caption{\textbf{Distributions of predicted scores.} We compare \textcolor{ViCEcolor}{\ourmethod}'s predictions to the \textcolor{GTcolor}{ground truth} using LLaVA-Video-7B model as the LMM.}
    \vspace{-10pt}
    \label{fig:kde}
\end{figure}

Tab.~\ref{tab:convs_main} and Tab.~\ref{tab:stepdiff_main} report the results obtained on ConViS-Bench~\cite{liberatori2025convisbench} and StepDiff~\cite{nagarajan2024step} respectively.
\ourmethod consistently outperforms the underlying LMMs across both datasets, yielding substantial gains on the \textit{zero-shot} baseline of up to $25.5\%$ on ConViS-Bench and $23.3\%$ on StepDiff on average.
This highlights the effectiveness of \ourmethod in leveraging LMMs within their preferred single-video input setting, while still enabling reliable cross-video comparison in a zero-shot regime.

\ourmethod elevates compact models to performance tiers typically reserved for much larger counterparts: a 2B model equipped with our framework consistently outperforms standard 8B baselines (\eg, $+11.9\%$ on average, Tab.~\ref{tab:convs_main}); while an 8B model surpasses even 32B baselines (\eg, $+6.8\%$ on average, Tab.~\ref{tab:convs_main}).
This shows that test-time inference is a compelling and resource-efficient alternative to parameter scaling, delivering higher accuracy at a fraction of the GPU memory cost.

Fig.~\ref{fig:box_plots} shows the score distribution of predictions made by the \textit{zero-shot} LMM and \ourmethod, as well as the ground truth scores. 
LLaVA-OV-7B produces generally high similarity scores across all concepts, with no scores falling below 3 for \textit{location}.
Applying \ourmethod mitigates this bias, yielding a wider distribution that better matches the ground truth.
For the Qwen3-VL models, the \textit{zero-shot} baseline consistently predicts low scores on ConViS-Bench, while skewing toward very high scores on StepDiff, whereas \ourmethod consistently produces distributions that better reflect the ground truth, including both low and high extremes.
For LLaVA-Video-7B on StepDiff, the \textit{zero-shot} baseline produces scores heavily peaked at 4; \ourmethod again yields more diverse, better-calibrated distributions.

Fig.~\ref{fig:kde} shows the distributions of \ourmethod's predicted scores against the ground truth using LLaVA-Video-7B across both datasets. 
On ConViS-Bench, \ourmethod's predictions are more closely aligned with the ground truth for the concepts of \textit{main action}, \textit{location}, and \textit{main objects}. 
In contrast, on StepDiff, the model performs better on the more object-centric concepts of \textit{tools/equipment} and \textit{ingredients}.
We attribute the weaker action-related performance on StepDiff to its finer granularity and the fact that its \textit{action} concept encompasses not only the action itself but also procedural step ordering.

\noindent\textbf{Qualitative results.}
Fig.~\ref{fig:qualitatives} presents a qualitative example, with additional ones in Appendix~\ref{sec:qualitatives_appendix}. 
The concept-conditioned descriptions effectively isolate the target concept, and the reasoning traces provide evidence that it influences the final score, yielding assessments that align more closely with ground-truth preferences than the \textit{zero-shot} baseline.

\begin{figure*}[h]
    \centering
    \includegraphics[width=\linewidth]{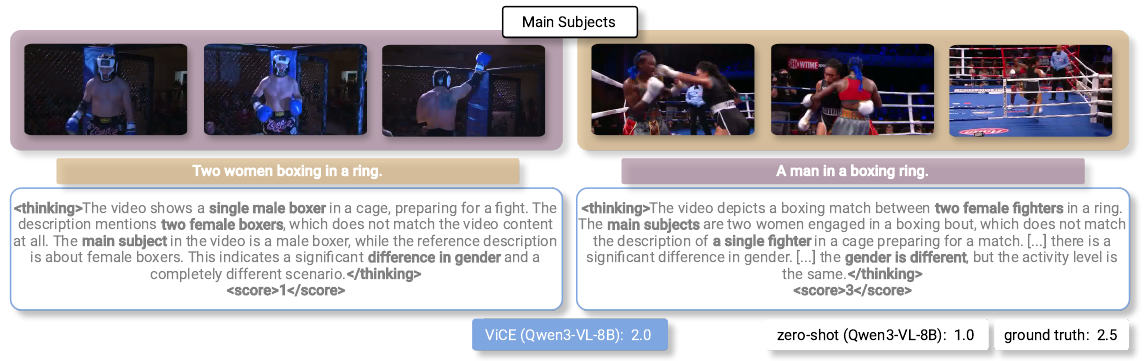}
    \caption{Example of a video pair with the captions and reasoning traces generated when conditioning on the target concept. Compared to the \textit{zero-shot} baseline, \ourmethod produces scores that more closely align with the ground truth.}
    \label{fig:qualitatives}
\end{figure*}

\begin{table*}[t!]
\def\arraystretch{1.0}
\setlength{\tabcolsep}{2pt}
\centering
\scriptsize
\resizebox{\textwidth}{!}{%
\begin{NiceTabular}{l@{\hskip15pt}c@{\hskip20pt}c@{\hskip19pt}c@{\hskip30pt}c@{\hskip19pt}c@{\hskip18pt}c@{\hskip 0pt}l}
\CodeBefore
   \rowcolors{7-8, 7-6, 13-14, 19-20, 25-26, 30-31}{sota}{}
   \rowcolors{2-3, 8-9, 14-15, 20-21, 26-27}{}{basemodel}
\Body
Model &
\textbf{\texttt{Main Action}} &
\textbf{\texttt{Main Subjects}} &
\textbf{\texttt{Main Objects}} &
\textbf{\texttt{Location}} &
\textbf{\texttt{Order of Actions}} &
Avg. \\
\midrule
\multicolumn{7}{l}{LLaVA-OV-7B~\cite{li2025llavaonevision}} \\
\ \ \blind & 53.4 & 38.5 & 42.9 & 47.0 & 8.1 & 38.0 \\
\plusours w/o \textit{CoT} & 48.1 & 48.9 & 46.7 & 57.2 & 20.2 & 44.2 \\
\plusours single branch & 47.6 & 48.4 & 43.0 & 58.4 & \textbf{25.0} & 44.5 \\
\plusours ($M=1$) & 48.2 & 49.2 & 47.1 & 60.8 & 40.3 & \textbf{49.1} \\
\plusours & \textbf{54.4} & \textbf{55.9} & \textbf{50.0} & \textbf{64.8} & 19.2 & 48.9 \\
\midrule
\multicolumn{7}{l}{LLaVA-Video-7B~\cite{zhang2025llavavideo}} \\
\ \ \blind & 53.3 & 40.2 & 44.4 & 48.7 & 42.2 & 45.8 \\
\plusours w/o \textit{CoT} & 57.4 & 50.1 & 54.2 & 56.7 & 49.7  & 53.6 \\
\plusours single branch & 56.2 & 48.6 & 52.0 & 59.1 & 50.4 & 53.3 \\
\plusours ($M=1$) & 56.6 & 49.4 & 57.3 & 61.3 & 52.0 & 55.3 \\
\plusours & \textbf{63.8} & \textbf{56.0} & \textbf{60.1} & \textbf{64.7} & \textbf{57.9} & \textbf{60.5} \\
\midrule
\multicolumn{7}{l}{Qwen3-VL-2B~\cite{bai2025qwen3}} \\
\ \ \blind & 47.5 & 36.2 & 42.6 & 41.3 & 44.3 & 42.4 \\
\plusours w/o \textit{CoT} & 50.0 & 41.5 & 53.6 & 61.5 & 50.3 & 51.4 \\
\plusours single branch & 52.8 & 41.3 & 50.5 & 55.4 & 44.5 & 48.9 \\
\plusours ($M=1$)  & 51.3 & 46.4 & 54.9 & 58.5 & 43.7 & 51.0 \\
\plusours   & \textbf{58.4} & \textbf{50.7} & \textbf{60.0} & \textbf{61.8} & \textbf{52.3} & \textbf{56.3} \\
\midrule
\multicolumn{7}{l}{Qwen3-VL-8B~\cite{bai2025qwen3}} \\
\ \ \blind & 54.8 & 39.3 & 51.0 & 45.8 & 45.1 & 47.2 \\
\plusours w/o \textit{CoT} & 59.0 & 46.2 & 58.3 & 68.7 & 59.2 & 58.3 \\
\plusours single branch & 55.8 & 42.8 & 49.9 & 62.0 & 52.3 & 52.6 \\
\plusours ($M=1$) & 61.6 & 45.2 & 57.9 & 67.5 & 58.7 & 58.2 \\

\plusours   & \textbf{64.6} & \textbf{49.6} & \textbf{61.4} & \textbf{69.8} & \textbf{62.7} & \textbf{61.6} \\

\midrule
\multicolumn{7}{l}{Qwen3-VL-32B~\cite{bai2025qwen3}} \\
\ \ \blind & 56.1 & 42.8 & 52.3 & 53.7 & 54.3 & 51.8 \\
\plusours w/o \textit{CoT} & 62.6 & 52.1 & 58.2 & 68.9 & 61.3 & 60.6 \\
\plusours single branch & 63.3 & 55.4 & 55.1 & 63.6 & 62.5 & 60.0 \\
\plusours ($M=1$) & 66.2 & 56.3 & 60.9 & 69.4 & 60.5 & 62.7 \\
\plusours   & \textbf{69.4} & \textbf{61.7} & \textbf{61.4} & \textbf{70.2} & \textbf{65.3} & \textbf{65.6} \\

\end{NiceTabular}
}
\caption{\textbf{Ablation Studies on ConViS-Bench}~\cite{liberatori2025convisbench}. We report Spearman’s $\rho$ ($\times100$).
}
\label{tab:ablation_0}
\end{table*}

\section{Ablation Studies}
\label{sec:ablations}

We present ablation studies on the contribution of each component of \ourmethod in Tab.~\ref{tab:ablation_0}.

\noindent\textbf{Blind.}
To isolate the contribution of visual information during the entailment step, we introduce a \textit{Blind} variant that removes the video–caption entailment entirely and replaces it with a purely text-based large language model (LLM). 
For each video pair, we provide the two concept-conditioned captions generated via Eq.~\eqref{eq:caption_gen} to the LLM and prompt it to predict a similarity score based solely on the text. We adopt LLaMA3-8B-Instruct~\cite{grattafiori2024llama} for this purpose.
Across all LMMs, the \textit{Blind} variant consistently yields lower performance than \ourmethod, indicating that visual inputs provide complementary information that is not necessarily captured by caption text alone.

\noindent\textbf{Effect of dual-branch design.}
The core idea of \ourmethod lies in its dual-path architecture. 
For this experiment, we caption one video only and entail it against the other one, having a \textit{single-branch} design.
Results show that while each branch contributes to the final prediction, a single branch alone is not sufficient to match \ourmethod's performance, showing the complementary nature of the branches and the effectiveness of their joint use.

\noindent\textbf{Effect of repeated sampling.}
\ourmethod generates multiple outputs within each branch, effectively exploring a richer set of reasoning paths. 
This produces similarity estimates that better align with human judgments, as different reasoning traces can capture complementary aspects of a concept. 
We show that a single-CoT approach ($M=1$) underperforms \ourmethod, highlighting the benefit of aggregating multiple reasoning processes for robustness.

\noindent\textbf{Effect of reasoning traces.}
\ourmethod produces concept-conditioned similarity scores along with corresponding reasoning traces. 
These traces add interpretability by revealing the model's decision process, while also improving score quality.
When we disable reasoning generation and prompt \ourmethod to output the final similarity score directly in the caption entailment step, performance drops by up to $6.9\%$ on average. 
This demonstrates that generating explicit reasoning traces encourages more structured and coherent representations, leading to more accurate similarity estimation.

\section{Conclusions}
\label{sec:conclusions}

In this work, we addressed concept-based video similarity, a task where existing LMMs struggle due to their sensitivity to video ordering.
We introduced \ourmethod, a novel test-time inference method based on a dual-branch, caption-then-entail approach, which additionally produces human-readable explanations that summarize 
the reasoning behind its predictions.
Experiments on two benchmarks show consistent improvements in both accuracy and interpretability over baselines that process video pairs jointly.
As an additional contribution to the community, we release 
human-curated concept-level similarity annotations for StepDiff video pairs, 
providing a new benchmark resource for concept-based video understanding.

\newpage 

\section{Limitations}
\label{sec:limitations}
Being a test-time inference method, \ourmethod's performance depends on the capabilities of the underlying LMM, and may suffer if the model cannot properly explain the concept of interest. 
However, this same property makes \ourmethod  flexible: without requiring concept-specific fine-tuning, it can be directly applied to any concept expressible in natural language. As LMMs advance, we expect \ourmethod's performance to improve correspondingly.

Additionally, \ourmethod requires more 
forward passes than a standard zero-shot baseline: two captioning operations and $2M$ 
entailment operations per video pair. 
However, the $M$ reasoning traces within each branch are generated in parallel 
via batched sampling, and the two branches can be parallelized as well, reducing 
the number of sequential forward passes to just two.
This test-time compute 
overhead enables smaller models to outperform larger ones, as discussed 
in Sec.~\ref{sec:experiments}.

Finally, \ourmethod is not designed for long videos, as concept-conditioned 
captioning may fail to capture all relevant moments under sparse frame sampling. 
We note, however, that cross-video understanding remains an open challenge even 
for short videos, and we see long-video comparison as a natural direction for 
future work as the field matures.

\section*{Acknowledgments}
This work was supported by the EU Horizon projects IAMI (No. 101168272), and ELLIOT (No. 101214398). We also thank the Deep Learning Lab of the ProM Facility for the GPU time.

\bibliography{main}

\clearpage
\appendix

\section{StepDiff Annotation}
\label{sec:stepdiff}

The original StepDiff~\cite{nagarajan2024step} dataset features video pairs annotated with captions that describe differences across five main categories, primarily focusing on textual descriptions of these differences. 
It also includes similarity scores for some categories.
However, the annotations are only partially complete: most video pairs have captions for fewer than five categories, and many similarity scores are either missing or reported as zero, which is outside the valid range. 
To enable more consistent and comprehensive evaluation for the \ourtask task, we annotated all video pairs using a clear scoring rubric. 
Each pair received similarity scores from 1 to 5 in all categories, ensuring complete and uniform coverage.

Human annotation was performed via Prolific~\cite{douglas2023data} and annotators were compensated at a rate of \pounds 8.50 per hour.
Since the videos in StepDiff are originally identified via YouTube IDs, some videos have been taken down or made private. 
Consequently, our reannotated dataset consists of the subset of videos that remain publicly available. 
In order to have video pairs with sufficient temporal structure, we kept those with videos longer than $10$ seconds.
The corresponding video IDs for this subset will be disclosed to facilitate reproducibility.

\section{Implementation Details}
\label{sec:impl_det}
All models are evaluated using $16$ frames per
video. 
The number of reasoning traces per branch is set to $M=10$. 
We use LLaMA3-8B-Instruct~\cite{grattafiori2024llama} as $f_{\texttt{LLM}}$ to generate explanations from reasoning traces in Sec.~\ref{sec:explain}.
When sampling multiple reasoning traces, we employ stochastic decoding with a temperature of $0.7$ and nucleus sampling (top-p=$0.9$).
For all other generations, we default to greedy decoding.
All experiments were conducted on a single NVIDIA A100 GPU.

\section{User Study}
\label{sec:user_study}
\begin{figure}
    \centering
    \includegraphics[width=\linewidth]{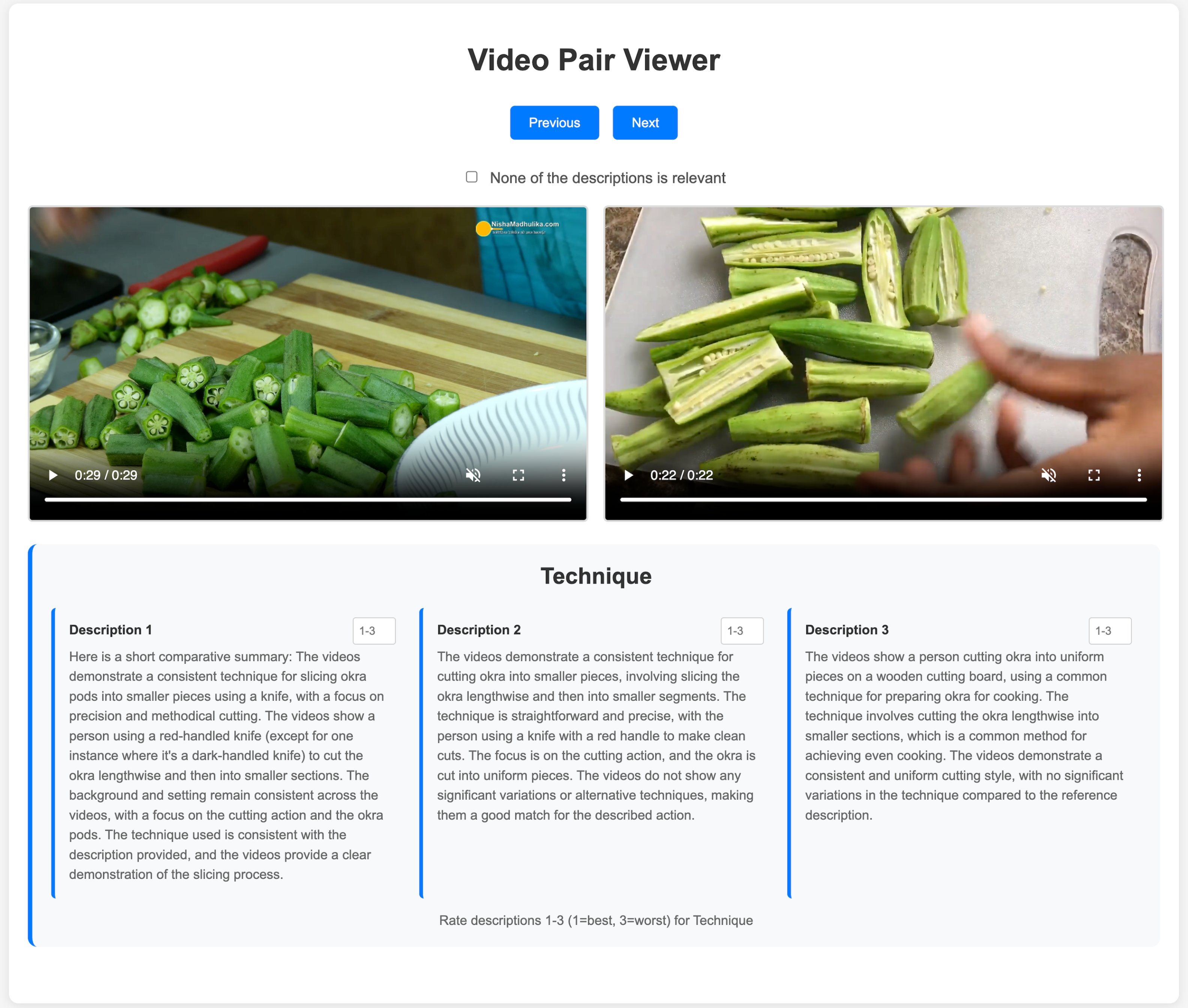}
    \caption{\textbf{User Study Interface:} Annotators are asked to rank the descriptions or select the checkbox if none of them apply.}
    \label{fig:interface}
\end{figure}

We conducted a user study to assess the quality of the explanations generated by \ourmethod (\cref{sec:explain}).
We randomly selected a total of 30 video pairs, drawn in equal proportion from ConViS-Bench~\cite{liberatori2025convisbench} and StepDiff~\cite{nagarajan2024step}.
For each video pair, we gathered explanations from three models~\cite{zhang2025llavavideo,li2025llavaonevision,bai2025qwen2}, generated as defined in \cref{sec:explain}.
We recruited five independent human raters, who were asked to rank the three explanations for each video pair from best to worst based on how well they described information relevant to the video pair content.
The three explanations for a given video pair were shown in randomized order to mitigate position bias.
Raters were instructed to focus solely on relevance and clarity, without considering model identity or stylistic preferences.
The interface used for the study is shown in Fig.~\ref{fig:interface}.

We find that in $97\%$ of cases, at least one explanation was considered relevant by the raters, highlighting the robustness of \ourmethod across different underlying models.
Tab.~\ref{tab:explanation_ranking} reports how often each model's explanation was ranked 1st, 2nd, or 3rd among the provided options, showing that models are broadly comparable, with Qwen3-VL-8B marginally outperforming the others, a trend consistent with the quantitative results in Tab.~\ref{tab:convs_main} and Tab.~\ref{tab:stepdiff_main}.

\begin{figure*}[t!]
    \centering
    \includegraphics[width=\linewidth]{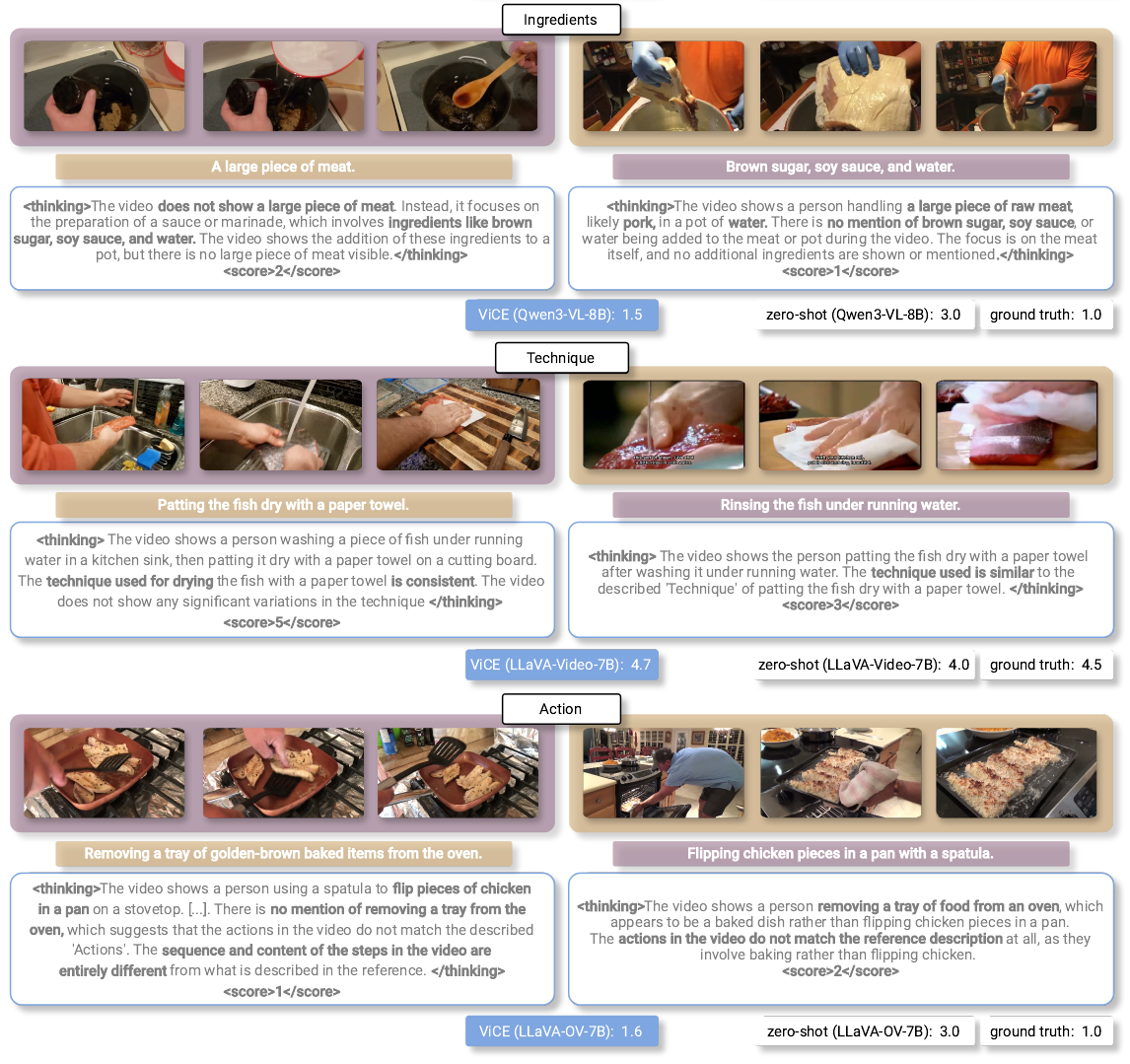}
    \caption{\textbf{Qualitative results of \ourmethod:} 
    We present three example video pairs for two distinct concepts and datasets~\cite{liberatori2025convisbench,nagarajan2024step},
    along with the captions and reasoning traces generated when conditioning on those concepts. 
    Compared to the \textit{zero-shot} baseline, \ourmethod produces scores that more closely align with the ground truth ones.}
    \label{fig:qualitatives_appendix}
\end{figure*}

\begin{table}[t]
\centering
\resizebox{\columnwidth}{!}{%
\begin{tabular}{lccc}
Model & 1$^{\text{st}}$(\%) & 2$^{\text{nd}}$(\%) & 3$^{\text{rd}}$(\%) \\
\midrule
Qwen3-VL-8B~\cite{bai2025qwen3} & 38.34 & 24.87 & 36.79 \\
LLaVA-Video-7B~\cite{zhang2025llavavideo} & 31.61 & 42.49 & 25.90 \\
LLaVA-OV-7B~\cite{li2025llavaonevision} & 29.53 & 33.16 & 37.31 \\
\end{tabular}}
\caption{\textbf{User Study Results:} We report the percentage of times each model's explanation generated by \ourmethod with different LMMs (see Sec.~\ref{sec:explain}) was ranked 1st, 2nd, or 3rd.}
\label{tab:explanation_ranking}
\end{table}

\begin{table}[t]
\centering
\resizebox{\columnwidth}{!}{%
\begin{tabular}{lcc}
Method & Peak GPU Memory (GiB) & Time / Prediction (s) \\
\midrule
Baseline (zero-shot) & 17.5 & 2.5 \\
VICE & 22.4 & 5.8 \\
\end{tabular}}
\caption{\textbf{Computational Cost:} Peak GPU memory and wall-clock time per prediction for the zero-shot baseline and VICE using Qwen3-VL-8B on a single NVIDIA A100 GPU.}
\label{tab:computation}
\end{table}

\section{Runtime and Memory Analysis}
VICE requires more forward passes than a direct video-pair comparison. However, most of these forwards (the $M$ sampled entailment traces per branch) can be batched rather than run sequentially, substantially limiting the added latency in practice.

To quantify this, we profile Qwen3-VL-8B on a single NVIDIA A100 GPU, measuring peak GPU memory and wall-clock time per prediction. Considering an 8B model, whose weights alone require 16.97 GiB (a fixed cost common to both methods), we report the average peak memory usage and inference time in Tab.~\ref{tab:computation}.
Despite the additional steps, VICE incurs only a moderate computational overhead.

\section{Qualitative Results}
\label{sec:qualitatives_appendix}

Fig.~\ref{fig:qualitatives_appendix} presents additional qualitative results complementing the example shown in Fig.~\ref{fig:qualitatives}.
This appendix is further accompanied by qualitative 
results in video format, accessible by opening \texttt{index.html} in the 
provided zipped folder.

\section{Licenses}

The datasets~\cite{liberatori2025convisbench, nagarajan2024step} used in our experiments are released under Creative Commons 
Attribution-NonCommercial 4.0 International (CC BY-NC 4.0) license.
The LMMs used in our experiments are all publicly available, released under the Apache 2.0 license.

\section{Prompts}
\label{sec:prompts}
Our proposed \ourmethod consists of two main steps, each using a dedicated prompt.

In the first concept-conditioned captioning step in Eq.~\eqref{eq:caption_gen}, we prompt the LMM to generate descriptions of the input video that are explicitly focused on a given concept.
In doing so, we provide guiding examples for each concept.

\begin{widebox}{Concept-conditioned captioning prompt}
You are given a video. Generate a caption that focuses specifically on the \textbf{\texttt{\{concept\}}} of the video.\\
Describe ONLY the \textbf{\texttt{\{concept\}}} and DO NOT include details about other aspects unless they are directly relevant to the \textbf{\texttt{\{concept\}}}.\\
Keep the caption natural, concise, and directly related to the \textbf{\texttt{\{concept\}}}.\\

Examples:\\
\textbf{\texttt{\{\textcolor{olive}{examples}[concept]\}}}\\

Here is the video. Generate a caption that describes its \textbf{\texttt{\{concept\}}} clearly and naturally.\\
\end{widebox}

\begin{widebox}{Video-caption entailment prompt}
You are given a text description and a video. \\
Your task is to evaluate how well the video matches the description with respect to the \textbf{\texttt{\{concept\}}}.\\[2pt]

Use the following rubric to determine your score:\\

\textbf{\texttt{\{\textcolor{pink}{rubric}[concept]\}}} \\[2pt]

Follow the format strictly:

- Inside \textbf{\texttt{<thinking> </thinking>}}, explain how the \textbf{\texttt{\{concept\}}} appears in the video and how it relates to the provided description, referencing the rubric levels where relevant.\\
- Then, output a single similarity score from 1 (least similar) to 5 (most similar) inside \textbf{\texttt{<score> </score>}}, reflecting how closely the video aligns with the description regarding the \textbf{\texttt{\{concept\}}}.\\

Format example:\\

\textbf{\texttt{<thinking>}}\\
Analysis of how the video matches the described \textbf{\texttt{\{concept\}}}.\\
\textbf{\texttt{</thinking>}}\\
\textbf{\texttt{<score>} }Predicted score \textbf{\texttt{</score>}}\\

Here is a video and a text description. Evaluate how well the video matches the description based on its \textbf{\texttt{\{concept\}}}, following the specified format with \textbf{\texttt{<thinking>}} and \textbf{\texttt{<score>}}.

\end{widebox}

\begin{widebox}[coltitle=olive]{Examples}
\texttt{\textbf{Main Subjects:}} \\
- A young boy with curly hair.\\
- An elderly man with glasses and a beard.\\[2pt]

\texttt{\textbf{Main Objects:}} \\
- A red bicycle leaning against a wall.\\
- A stack of books on a wooden table.\\[2pt]

\texttt{\textbf{Location:}} \\
- A bustling city street lined with shops.\\
- A quiet park with tall trees.\\[2pt]

\texttt{\textbf{Main Action:}} \\
- People are tossing a ball back and forth.\\
- A cat jumps onto the counter.\\[2pt]

\texttt{\textbf{Order Of Actions:}} \\
- The chef chops vegetables, then sautés them in a pan.\\
- She opens the door, walks outside, and waves goodbye.\\[2pt]

\texttt{\textbf{Tools/Equipment:}} \\
- A sharp chef’s knife beside a wooden cutting board.\\
- A stainless steel saucepan simmering on the stove.\\[2pt]

\texttt{\textbf{Actions:}} \\
- Whisking the batter until smooth.\\
- Searing the steak on a hot pan.\\[2pt]

\texttt{\textbf{Visuals:}} \\
- A dark brown sauce coating the meat.\\
- Crispy golden-brown eggs on the plate.\\[2pt]

\texttt{\textbf{Ingredients:}} \\
- Fresh tomatoes, basil, and mozzarella on the counter.\\
- Flour, sugar, and eggs arranged in small bowls.\\[2pt]

\texttt{\textbf{Technique:}} \\
- Using gentle folding motions to mix the batter.\\
- Spreading the sauce evenly across the plate with the back of a spoon.\\
\end{widebox}

In the second video-caption entailment step in Eq.~\eqref{eq:ent_step}, we prompt the LMM to assess the semantic alignment between a video and a textual description with respect to a given concept. 
We guide this estimation using a rubric, identical to the one given to human raters.

\begin{widebox}[coltitle=pink]{Rubric ConViS-Bench}
\small
\textbf{\texttt{Main Action:}} \\
Score $\mid$ Description \\
1 $\mid$ Completely different types of action (e.g., cooking and sports domains)\\
2 $\mid$ Actions are different but related (e.g., both sports)\\
3 $\mid$ Actions are similar with significant differences in how they are carried out\\
4 $\mid$ Actions are similar with slight differences in how they are carried out\\
5 $\mid$ No difference in the main action performed\\

\textbf{\texttt{Main Subjects:}} \\
Score $\mid$ Description\\
1 $\mid$ Totally different cardinality, gender, and age (e.g., a man and a group of women)\\
2 $\mid$ Different cardinality but some similar in gender or age (e.g., a man and a group of men)\\
3 $\mid$ Same cardinality but significant difference in age or gender (e.g., a man and a boy)\\
4 $\mid$ Same cardinality but minor difference in age (e.g., a young male and a boy)\\
5 $\mid$ Extremely similar (e.g., two young women in both)\\

\textbf{\texttt{Location:}}\\
Score $\mid$ Description\\
1 $\mid$ Completely different places\\
2 $\mid$ Different but similar type of place (e.g., indoor and outdoor soccer field)\\
3 $\mid$ Similar locations still with significant differences\\
4 $\mid$ Similar locations with slight differences\\
5 $\mid$ Extremely similar (e.g., rooms with a similar size and layout)\\

\textbf{\texttt{Order Of Actions:}} \\
Score $\mid$ Description\\
1 $\mid$ Completely different, or no sub-actions are in common\\
2 $\mid$ The order of actions has significant changes, and some missing steps\\
3 $\mid$ Some actions are reordered, with a few steps missing, but key steps remain intact\\
4 $\mid$ Minor changes in the order of actions\\
5 $\mid$ Actions occur in the same order, without missing steps\\

\textbf{\texttt{Main Objects:}} \\
Score $\mid$ Description\\
1 $\mid$ Main objects are entirely different\\
2 $\mid$ A few main objects are shared, but most are different\\
3 $\mid$ Most main objects are similar, with some not in common\\
4 $\mid$ Minor differences in main objects (e.g., color or shape)\\
5 $\mid$ Main objects are the same or extremely similar\\
\\
\end{widebox}

\begin{widebox}[coltitle=pink]{Rubric StepDiff}
\small
\textbf{\texttt{Ingredients:}} \\
Score $\mid$ Description\\
1 $\mid$ Completely different main ingredients (e.g., pork vs fish)\\
2 $\mid$ Major ingredients differ, but some elements remain the same\\
3 $\mid$ Several minor ingredients are added or skipped (3 or more instances)\\
4 $\mid$ Only one or two minor ingredients differ (added, skipped, or substituted)\\
5 $\mid$ No difference in ingredients used\\

\textbf{\texttt{Tools/Equipment:}} \\
Score $\mid$ Description\\
1 $\mid$ Completely different tools or appliances are used\\
2 $\mid$ Multiple substitutions or material changes (4 or more) that serve the same purpose\\
3 $\mid$ A few tools (2–3) are substituted or differ in material but serve the same function\\
4 $\mid$ Minor differences such as material (e.g., metal vs wood), or type within the same category (e.g., non-stick vs ordinary pan)\\
5 $\mid$ No difference in tools or equipment used except for minor aspects like color or size\\

\textbf{\texttt{Actions:}}\\
Score $\mid$ Description\\
1 $\mid$ Sequence and content of steps are entirely different\\
2 $\mid$ Sequence differs with two to three steps added or skipped\\
3 $\mid$ Some steps are added, skipped, or slightly changed but the main process remains\\
4 $\mid$ Order of steps is slightly altered but all actions are present\\
5 $\mid$ All steps occur in the same sequence with no differences\\

\textbf{\texttt{Technique:}}\\
Score $\mid$ Description\\
1 $\mid$ Actions are different and techniques are not comparable\\
2 $\mid$ Techniques are different (e.g., boiling vs steaming)\\
3 $\mid$ Two to three noticeable variations in cooking or preparation techniques\\
4 $\mid$ Slight procedural variation (e.g., slicing vs dicing) with similar outcome\\
5 $\mid$ No difference in techniques used\\

\textbf{\texttt{Visuals:}}\\
Score $\mid$ Description\\
1 $\mid$ Completely different appearance, with major differences in color or consistency (e.g., brown vs white)\\
2 $\mid$ Significant difference in size or portioning of dishes\\
3 $\mid$ Minor visual differences such as slightly darker or crisper elements\\
4 $\mid$ Very small variation in color or texture affecting only a small portion or detail\\
5 $\mid$ No visible difference in final appearance\\
\end{widebox}

\end{document}